\documentclass{article}

\usepackage{hyperref}

\usepackage{microtype}
\usepackage{graphicx}
\usepackage{booktabs} 

\usepackage[preprint]{icml2026}

\usepackage{amsmath}
\usepackage{mathtools}
\usepackage{amsthm}
\usepackage{colortbl}

\usepackage[capitalize,noabbrev]{cleveref}

\theoremstyle{plain}

\theoremstyle{definition}

\theoremstyle{remark}

\usepackage[textsize=tiny]{todonotes}

\usepackage{layouts}
\usepackage{url}
\usepackage{subfigure}
\usepackage{wrapfig}
\usepackage{xspace}

\usepackage{multirow}
\usepackage{xcolor}
\usepackage{tabularray}
\usepackage{algorithm}

\makeatletter
\@ifundefined{algorithmic}{}{%

}
\@ifundefined{State}{}{%
  \let\State\relax
}
\makeatother

\usepackage{algpseudocode}

\usepackage{wrapfig,booktabs,multirow,makecell,calc}
\usepackage{pifont}
\usepackage{enumitem}
\usepackage{amssymb}

\newcommand{\modelname}{{\fontfamily{lmtt}\selectfont \textbf{PLUS}}\xspace}

\icmltitlerunning{Planning-Augmented Sampling with Early Guidance for High-Reward Discovery}

\begin{document}

\twocolumn[
  \icmltitle{Planning-Augmented Sampling with Early Guidance for High-Reward Discovery}



  \icmlsetsymbol{equal}{*}

  \begin{icmlauthorlist}
    \icmlauthor{Rui Zhu}{ustc}
    \icmlauthor{Yudong Zhang}{ustc}
    \icmlauthor{Xuan Yu}{ustc}
    \icmlauthor{Chen Zhang}{ai-lab}
    \icmlauthor{Xu Wang}{ustc}
    \icmlauthor{Yang Wang}{ustc}
  \end{icmlauthorlist}

  \icmlaffiliation{ustc}{University of Science and Technology of China, Hefei, China}
  \icmlaffiliation{ai-lab}{Shanghai AI Laboratory, Shanghai, China}

  \icmlcorrespondingauthor{Yudong Zhang (Primary Contact)}{yudong.zhang@ustc.edu.cn}
  \icmlcorrespondingauthor{Yang Wang}{angyan@ustc.edu.cn}

  \icmlkeywords{Machine Learning, ICML}

  \vskip 0.3in
]

\printAffiliationsAndNotice{}  

\begin{abstract}
Generative Flow Networks (GFlowNets) enable structured generation with inherent diversity, but existing sampling strategies often rely on weak guided exploration, slowing early discovery of high-reward candidates. In tasks such as molecular design, rapid and consistent generation of high-reward solutions can outweigh faithful distribution matching.
We propose a planning-augmented framework in which Monte Carlo Tree Search using polynomial upper confidence bounds provides online value estimates, and a controllable soft-greedy mechanism integrates these planning signals into the GFlowNets forward policy. This design fosters early exploration of high-reward trajectories and gradually shifts to policy-driven exploitation as experience accumulates.
Empirical results show that our method accelerates early high-reward discovery, sustains top-quality sample generation, and preserves diversity across representative tasks.
All implementations are available at \hyperlink{https://github.com/ZRNB/PLUS}{https://github.com/ZRNB/PLUS}.
\end{abstract}

\section{Introduction}
Generative Flow Networks (GFlowNets)~\citep{bengio2021flow,gao2022sample,bengio2023gflownet} have recently emerged as a powerful tool for generating diverse high-quality candidates by learning to sample from a reward-proportional distribution. This property makes GFlowNets particularly attractive for a wide range of structured generation tasks. 
For example,
~\citet{deleu2022bayesian} and~\citet{nishikawa2022bayesian} employ GFlowNets to model posterior distributions over discrete compositional structures such as Bayesian networks. 
\citet{liu2023gflowout} utilizes GFlowNets for sampling modular subnetworks, improving model generalization under distributional shifts.
However, vanilla GFlowNets often struggle to efficiently discover high-reward samples in complex environments. While their inherent exploratory nature enhances diversity, it may lead to excessive coverage of low-reward regions, particularly during early training when the sampling policy lacks guidance and relies on self-collected experience. 
As a result, high-reward regions may be discovered slowly, leading to suboptimal performance in sparse reward scenarios.

Broadly speaking, existing approaches to high-reward sampling can be categorized into two optimization paradigms with distinct objectives. The first paradigm focuses on approximating a target distribution by maximizing mode coverage, thereby indirectly revealing high-reward regions~\cite{madan2025towards, kim2025adaptive}. The second paradigm, by contrast, prioritizes the rapid discovery of high-reward candidates by explicitly biasing the sampling process toward greedier strategies~\cite{lau2024qgfn}. While both paradigms are valuable in different contexts, many real-world applications, such as molecular design, primarily seek to identify a set of high-reward candidates as early and as plentifully as possible. In such settings, efficient high-reward discovery is often more critical than faithful distributional approximation. \ding{96} This observation motivates the following question: 
\textbf{How can a model efficiently discover high-reward regions early in training, while sustaining high-reward generation as experience accumulates?}

\begin{figure*}
    \centering
    \includegraphics[width=\linewidth]{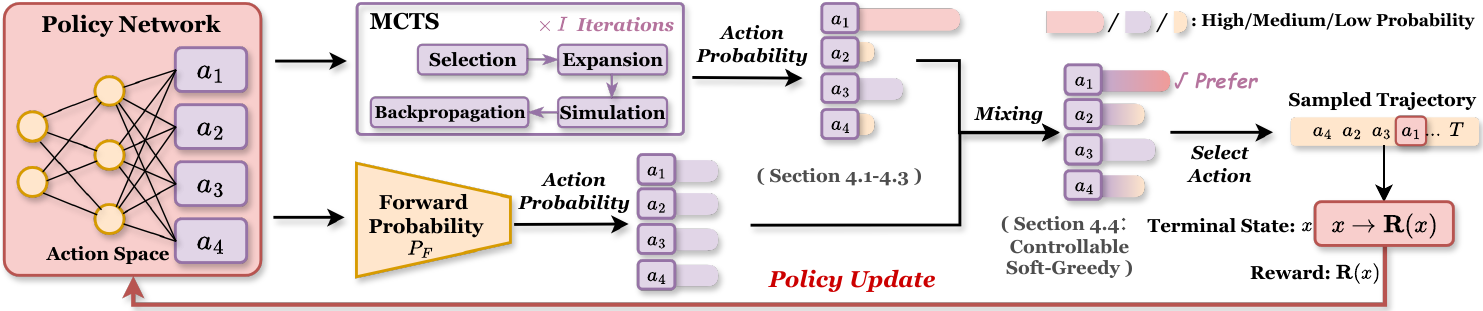}
    \caption{Framework overview. Candidate actions from the policy network are evaluated via MCTS iterations to produce action probabilities, which are mixed with forward probabilities to generate trajectories toward terminal states $x$ with rewards $\mathcal{R}(x)$.}
    \label{fig: introduction}
    \vspace{-0.3cm}
\end{figure*}

Motivated by this perspective, we explicitly adopt a greedier sampling strategy aimed at efficient high-reward discovery, rather than emphasizing strict distributional approximation. To this end, we propose a \textbf{\underline{PL}}anning-a\textbf{\underline{U}}gmented \textbf{\underline{S}}ampling framework (\modelname) that integrates Monte Carlo Tree Search (MCTS) with Polynomial Upper Confidence Trees (PUCT)~\citep{kocsis2006bandit} and a tunable $\alpha$-greedy sampling mechanism. As illustrated in Fig.~\ref{fig: introduction}, the final action distribution is obtained by blending the action probabilities estimated by MCTS with those predicted by the GFlowNets forward policy. The MCTS-derived probabilities encode the estimated utility of state–action pairs, and their incorporation biases sampling toward actions with higher expected returns. Through this controlled mixing, the framework enables explicitly greedy yet resource-efficient sampling, facilitating rapid discovery of high-reward candidates.
Our main \textbf{contributions} are as follows:
\vspace{-0.3cm}
\begin{itemize}[leftmargin=1.5em,itemsep=-0.1em]
\item[\ding{182}] We propose \modelname, a planning-augmented sampling framework that accelerates the discovery of high-reward candidate samples and increases their quantity, while preserving diversity.
\item[\ding{183}] We introduce a controllable soft-greedy strategy, allowing controllable prioritization of actions likely to yield high-reward states.
\item[\ding{184}] Empirical evaluation shows that \modelname significantly improves sampling efficiency and high-reward discovery, while preserving the diversity of generated solutions.
\end{itemize}
\vspace{-0.4cm}

\section{Related Work}
\vspace{-0.1cm}
\subsection{High-Reward Discovery in GFlowNets}
\vspace{-0.1cm}
GFlowNets provide a principled framework for structured generation via stochastic sampling~\citep{bengio2021flow}. Extensions span variational inference \citep{zimmermann2022variational}, credit assignment \citep{pan2023better}, multi-objective generation \citep{jain2022biological}, world modeling \citep{pan2023stochastic}, unsupervised learning \citep{pan2023pre}, bias reduction \citep{ma2024baking}, and links to energy-based and diffusion models \citep{zhang2023diffusion}. Nevertheless, efficiently discovering high-reward solutions in sparse-reward environments remains challenging.
Existing approaches fall into two paradigms. The first enhances distributional expressiveness or coverage to approximate the target distribution and indirectly reveal high-reward regions~\citep{madan2025towards, kim2025adaptive, boussif2025action}. While improving exploration, these methods do not explicitly address inefficient early-stage guidance.
The second paradigm directly biases sampling toward greedier behavior to prioritize high-reward solutions~\citep{lau2024qgfn}. Effective once value estimates stabilize, these methods remain limited in early exploration and may introduce persistent biases.
\vspace{-0.2cm}

\subsection{Planning-Based Guidance for Efficient Exploration}
\vspace{-0.1cm}
Planning and search methods have long enhanced exploration in sequential decision-making, particularly under sparse or delayed rewards. MCTS achieves a balance between exploration and exploitation via selective lookahead and has seen notable success in reinforcement learning \citep{silver2016mastering, silver2018general}. Polynomial Upper Confidence Trees (PUCT) further integrates value estimates with visitation statistics to guide exploration under uncertainty \citep{kocsis2006bandit}.
Recent work has drawn connections between GFlowNets and reinforcement learning, including value-based formulations such as QGFN \citep{lau2024qgfn}. More broadly, planning strategies have been applied to generative and combinatorial decision processes, often as the primary decision mechanism. In contrast, our approach leverages planning-derived signals to guide sampling, targeting inefficient early exploration while enabling a gradual shift toward exploitation during training.

\vspace{-0.2cm}
\section{Preliminary}
\vspace{-0.1cm}
\subsection{Generative Flow Networks}
\vspace{-0.1cm}
The generation process is modeled as a directed acyclic graph (DAG), where nodes correspond to states, and directed edges correspond to valid state transitions. A trajectory is defined as
$\tau = (s_0 \rightarrow s_1 \rightarrow \cdots \rightarrow s_n)$,
where \( s_n = x \) is a terminal state. We denote the set of all states by \( \mathcal{S} \), the set of terminal states by \( \mathcal{X} \), and the set of complete trajectories by \( \mathcal{T} \). An action is represented by a directed edge \( (s \rightarrow s') \). We are given a pointed DAG $\mathcal{G}=(\mathcal{S}, \mathbb{A)}$, where $\mathcal{S}$ is a finite set of vertices (\textit{states}), and $\mathbb{A} \subset \mathcal{S} \times \mathcal{S}$ is a set of directed edges (\textit{actions}). If $s \rightarrow s'$ is an action, we say $s$ is a parent of $s'$, and $s'$ is a child of $s$. We denote the parent set of $s$ as $\mathcal{P}(s)$, the child set of $s$ as $\mathcal{C}(s)$.
For a state $s$, let \( \mathcal{A}(s) \) denote the set of available outgoing actions from $s$; for any terminal state \( x \), \( \mathcal{A}(x) = \varnothing \). A trajectory flow is a nonnegative function $F: \mathcal{T} \rightarrow \mathbb{R}_{\geq 0}$. For any state $s$, define the state flow $F(s)=\sum_{s \in \tau} F(\tau)$, and, for any edges $s \rightarrow s'$, the edge flow
    \vspace{-0.2cm}
\begin{equation}
\small
    F(s \rightarrow s') = \sum_{\tau = (... s \rightarrow s' ...)} F(\tau).
    \label{eq: the definition of edge flow}
    \vspace{-0.2cm}
\end{equation}
As a consequence of this definition, the \textit{flow matching} constraint (incoming flow = outgoing flow) is satisfied for all states $s$ that are not initial or terminal:
\vspace{-0.1cm}
\begin{equation}
\small
    F(s) = \sum_{(s'' \rightarrow s) \in \mathcal{A}} F(s'' \rightarrow s) = \sum_{(s \rightarrow s') \in \mathcal{A}} F(s \rightarrow s').
    \label{eq: flow matching constraint}
    \vspace{-0.2cm}
\end{equation}
Anontrivial trajectory flow $F$ determines a distribution $P$ over trajectories,
\vspace{-0.2cm}
\begin{equation}
\small
    P(\tau) = \frac{1}{Z} F(\tau), \quad Z = F(s_0) = \sum_{\tau \in \mathcal{T}} F(\tau).
    \label{eq: P and Z definition}
    \vspace{-0.2cm}
\end{equation}
The forward and backward transition probabilities induced by the flows are given by
\vspace{-0.2cm}
\begin{equation}
\small
    P_F(s' \mid s) = \frac{F(s \rightarrow s')}{F(s)}, \,
    P_B(s \mid s') = \frac{F(s \rightarrow s')}{F(s')}.
    \label{eq: PF and PB definition}
    \vspace{-0.2cm}
\end{equation}
\textbf{\textit{Remark:}} If action $a$ is performed by state $s$ and it leads to state $s'$, since this action $a$ is unique, then $P_F(s' \mid s)$ is equivalent to $P_F(a \mid s)$.
Under the trajectory balance (TB) constraint~\citep{malkin2022trajectory}, for any complete trajectory $\tau$, the following condition holds:
\vspace{-0.2cm}
\begin{equation}
\small
Z \prod_{t=1}^{n} P_F(s_t \mid s_{t-1})
=
\mathcal{R}(x) \prod_{t=1}^{n} P_B(s_{t-1} \mid s_t).
\vspace{-0.2cm}
\end{equation}
A model with parameters $\theta$ outputs forward policy $P_{F}(\,\cdot\,|s;\theta)$, backward policy $P_{B}(\,\cdot\,|s;\theta)$, and a global scalar $Z_{\theta}\approx F(s_{0})$; together they induce an implicit Markovian flow $F_{\theta}$. For a trajectory $\tau$, define the TB loss:
\vspace{-0.2cm}
\begin{equation}
\small
\mathcal{L}_{\mathrm{TB}}(\tau)
=
\left(
\log
\frac{
Z_{\theta} \prod_{t=1}^{n} P_F(s_t \mid s_{t-1}; \theta)
}{
\mathcal{R}(x) \prod_{t=1}^{n} P_B(s_{t-1} \mid s_t; \theta)
}
\right)^2.
\label{eq: TB loss}
\vspace{-0.2cm}
\end{equation}

\vspace{-0.2cm}
\subsection{Monte Carlo Tree Search for GFlowNets}
\vspace{-0.1cm}
MCTS is a general search paradigm that iteratively refines action selection through \emph{selection}, \emph{expansion}, \emph{simulation}, and \emph{backpropagation}, guided by statistics accumulated during search. For each admissible action $a = (s \to s') \in \mathcal{G}$ at state $s$, MCTS maintains an empirical action-value $Q(a \mid s)$ and visit count $N(a \mid s)$, updated by backpropagating terminal rewards along sampled trajectories.
During rollouts, actions are chosen according to a policy balancing exploitation of high-value actions with exploration of less-visited alternatives. Integrating MCTS into GFlowNets involves associating each state–action pair with these statistics and updating them online during trajectory generation. Unlike standard GFlowNets relying solely on the forward policy $P_F$, the availability of $Q(a \mid s)$ and $N(a \mid s)$ provides a structured mechanism to bias sampling toward higher-reward regions while maintaining exploration of under-sampled areas.

\vspace{-0.3cm}
\subsection{Sampling Objective}
\vspace{-0.1cm}
We consider a sequential generation problem in which a structured object is constructed by applying a sequence of actions starting from an initial state $s_0$ and terminating at a final state $x$. Each terminal state $x$ is associated with a non-negative reward \( \mathcal{R}(x) \in \mathbb{R}^+ \), \textbf{rather than strictly approximating the reward-proportional distribution}, our goal is to \textbf{efficiently discover high-reward solutions as early and as plentifully as possible}, by intentionally introducing a controllable bias toward high-reward regions. 

\vspace{-0.28cm}
\section{Methodology}
\vspace{-0.1cm}
\paragraph{Overview.} To efficiently explore high-reward regions early while leveraging accumulated experience later, we equip GFlowNets with a targeted planning mechanism. Rather than sampling solely from the forward policy $P_F$, short, state-conditioned rollouts evaluate the long-term potential of available actions, prioritizing high-reward regions during early exploration while retaining the ability to exploit well-learned trajectories.

Formally, state $s$ undergo $I$ planning iterations, each comprising selection, expansion, simulation, and backpropagation. Each visited state instantiates a node storing action-value estimates $Q(a\mid s)$ and visit counts $N(a\mid s)$, capturing expected rewards and prior exploration. These estimates are iteratively refined throughout training.
After $I$ iterations, the updated $Q(a\mid s)$ values are integrated with $P_F$ via a tunable parameter $\alpha$, producing a normalized sampling distribution that balances greediness and exploration. An action is sampled from this distribution to transition to the next state $s'$, and the process repeats until a terminal state $x$ is reached, yielding a complete trajectory $\tau = (s_0 \to s_1 \to \cdots \to x)$.

\vspace{-0.2cm}
\subsection{Targeted Selection of High-Reward Regions}
\label{sec: Targeted Selection of high-reward Regions via PUCT}
\vspace{-0.1cm}
This stage identifies the state node most likely to yield high rewards during the internal search process. The selected trajectory is not executed in the environment; instead, it serves to guide the subsequent expansion, simulation, and backpropagation steps within each MCTS iteration. Actual action selection and state transition are performed only after completing the full set of $I$ MCTS iterations.
The selection stage traverses the state space starting from the current state $s$, with the explicit objective of balancing early-stage exploration of high-reward regions and later exploitation of accumulated experience. At each intermediate state $s$, we consider all outgoing actions $a \in \mathcal{A}(s)$ and maintain two statistics: the action-value estimate $Q(a\mid s)$ and the visit count $N(a\mid s)$.
Where all admissible actions $a \in \mathcal{A}(s)$ have been previously visited, with associated statistics $N(a\mid s)$ and $Q(a\mid s)$ available for every $a$, we define the cumulative visitation count at state $s$ as $N_{\text{tot}}(s) = \sum_{a' \in \mathcal{A}(s)} N(s, a')$. The PUCT scores are then computed as:
\vspace{-0.2cm}
\begin{equation}
\small
    \operatorname{PUCT}(a\mid s) = Q(a\mid s) + c_{\text{puct}} \cdot P_F(a \mid s) \cdot \frac{\sqrt{N_{\text{tot}}(s)}}{1 + N(a\mid s)}.
    \label{eq: puct_short}
    \vspace{-0.2cm}
\end{equation}
where $c_{puct}$ controls the trade-off between exploitation of high $Q(a\mid s)$ paths and exploration. Larger $c_{puct}$ values encourage more exploration—critical for uncovering high-reward regions early in training—while smaller values favor greedy exploitation, which helps sustain high-reward generation once informative experience has been accumulated.
To ensure numerical stability, we convert PUCT scores into a categorical distribution via softmax with max subtraction:
\vspace{-0.4cm}
\begin{align}
\small
    \tilde{v}_a = \exp\!\big(\operatorname{PUCT}(a\mid s) &- \max_{a} \operatorname{PUCT}(a\mid s)\big), \\
    p^{\text{PUCT}}(a\mid s) &= \frac{\tilde{v}_a}{\sum_{k} \tilde{v}_{a_k}}.
    \vspace{-0.2cm}
\end{align}
An action $a \sim \text{Categorical}({p^{\text{PUCT}}(a\mid s)}), {a \in \mathcal{A}(s)}$ is then selected, determining the next state $s'$ via the transition $s \xrightarrow{a} s'$. Selection terminates either when a terminal state $x$ is reached (at which point backpropagation begins) or when we encounter a state $s'$ with at least one unvisited action $(N(a\mid s')=0)$, ensuring that the search both refines high-reward regions and efficiently surfaces previously unexplored areas.
By explicitly incorporating both $Q(a\mid s)$, which encodes the cumulative reward information, and $N(a\mid s)$, which tracks prior exploration, the PUCT-guided selection mechanism dynamically prioritizes the discovery of high-reward trajectories early in training while naturally shifting toward exploitation of accumulated knowledge as the visit statistics become more informative. This design directly addresses the dual challenge articulated in the Introduction: enabling efficient early-stage exploration and sustained high-reward generation in later stages.

\vspace{-0.2cm}
\subsection{Simulation Strategy for Newly Expanded States}
\label{sec: Simulation Strategy for Newly Expanded States}
\vspace{-0.1cm}
Following the selection stage, the expansion stage initializes all previously unexplored actions originating from the selected state $s$. Specifically, for each unvisited action $a \in \mathcal{A}(s)$, we initialize its statistics by setting $N(a \mid s) = 0$ and $Q(a \mid s) = 0$.
The objective of the simulation stage is to estimate the reward attainable from a newly expanded state. Among the multiple child states produced during expansion, it is necessary to identify one state from which to simulate the subsequent sampling process until a terminal state $x$ is reached, thereby obtaining a corresponding reward $\mathcal{R}(x)$.
\ding{46} \textit{Which state should be regarded as the most suitable candidate for simulation?}
To address this question, we again employ the PUCT criterion defined in Eq.~\ref{eq: puct_short}, which balances action-value estimates $Q(a \mid s)$ against visit counts $N(a \mid s)$. However, for newly expanded states, all actions satisfy $Q(a \mid s) = 0$ and $N(a \mid s) = 0$. Under this condition, the PUCT formulation reduces to a comparison based solely on the magnitude of the forward policy prior $P_F$. Consequently, action selection at this stage is performed by sampling according to $P_F$.
Once an action is selected, the corresponding successor state $s'$ is obtained. Starting from $s'$, we simulate the remaining trajectory $(s' \rightarrow \cdots \rightarrow x)$ by following the forward policy $P_F$ until a terminal state $x$ is reached, at which point the reward $\mathcal{R}(x)$ is evaluated. This completes the simulation phase.

\vspace{-0.2cm}
\subsection{Accurate Action Value Estimation via Reward Propagation}
\label{sec: Reward Propagation for Accurate Action Value Estimation}
\vspace{-0.2cm}
During the backpropagation stage, the update processes of $Q(a\mid s)$ and $N(a\mid s)$ are of great significance, as they are relied upon for accuracy in all four stages of MCTS. Our method employs incremental updates for $Q(a\mid s)$ backpropagation, where each state's $Q(a\mid s)$ and $N(a\mid s)$ are updated as:
\begin{equation}
\small
    N(a\mid s) \leftarrow N(a\mid s) + 1,
    \label{eq: n_visit update}
    \vspace{-0.1cm}
\end{equation}
\begin{equation}
\small
    Q(a\mid s) \leftarrow Q(a\mid s)+\frac{ \mathcal{R}(x)-Q(a\mid s)}{N(a\mid s)}.
    \label{eq: Q-value update}
    \vspace{-0.1cm}
\end{equation}
The updates of $Q(a\mid s)$ and $N(a\mid s)$, defined in Eq.~\ref{eq: Q-value update} and Eq.~\ref{eq: n_visit update}, jointly balance exploration and exploitation.
The visit count $N(a\mid s)$ controls exploration: for each state-action pair encountered along the path from the selection stage to the expansion stage, $N(a\mid s)$ is incremented by one. This targeted credit assignment concentrates learning signals along productive pathways while adhering to the constraint that only visited trajectories can yield terminal rewards. As $N(a\mid s)$ increases, the exploration term in Eq.~\ref{eq: puct_short} decreases, reducing repeated exploration of the same region and alleviating the tendency to converge to local optima.
The evolution of $Q(a\mid s)$ governs the expected utility of an action; as the exploration process unfolds, these values increasingly approximate the true value of the action, thus biasing subsequent decisions toward state-action pairs that culminate in terminal states with higher rewards.
The two updates are complementary and jointly determine the selection behavior in Section~\ref{sec: Targeted Selection of high-reward Regions via PUCT}.
More details of the challenge in the backpropagation stage are provided in Section~\ref{sec: Backpropagation Challenge Details}.
\vspace{-0.2cm}
\subsection{Controllable Greedy Action Selection for High-Reward Sampling}
\label{sec: Controllable Greedy Action Selection for High-Reward Sampling}
\vspace{-0.2cm}
This stage constitutes the decision-making and action-execution phase, and is designed to realize a more greedy sampling strategy.
To seamlessly integrate the guidance of the forward policy $P_F$ with the informativeness encapsulated in $Q(a\mid s)$ for selecting an action $a$ at state $s$, we introduce a $\alpha$-greedy strategy in Eq.~(\ref{eq: categorical}) that dynamically modulates the relative contributions of the forward policy and the $Q(a\mid s)$-derived value distribution:
\vspace{-0.2cm}
\begin{equation}
\small
    p^{Q}_{i}=\frac{Q(a_i\mid s)- Q_{min}(a\mid s)+ \epsilon}{\sum_{k}\left(Q(a_k\mid s)- Q_{min}(a\mid s) + \epsilon\right)},
    \label{eq: pi}
    \vspace{-0.2cm}
\end{equation}
\begin{equation}
\small
    \mu \sim \operatorname{Categorical} \Big(\frac{(1-\alpha)\cdotp {P}_{F}+\alpha \cdotp p^{Q}_i}{{\left \| (1-\alpha)\cdotp {P}_{F}+\alpha \cdotp p^{Q}_i\right \|}_{1}} \Big).
    \label{eq: categorical}
    \vspace{-0.1cm}
\end{equation}
where $Q(a_i\mid s)$ denotes the value obtained by executing action $a_i$ from state $s$, and $Q_{\min}(a\mid s)$ is the minimal $Q(a\mid s)$ attainable among all admissible actions at $s$. The term $\epsilon$ is introduced to preclude division by zero. In the event that all $Q(a\mid s)$ values are identical, Eq.~(\ref{eq: pi}) degenerates to a uniform distribution.
Rather than adopting a greedy strategy that selects the action with maximal $Q(a\mid s)$, we construct a probability distribution by linearly rescaling the $Q(a\mid s)$ values to guarantee non-negativity and proper normalization. Directly choosing the highest-value action can precipitate premature convergence, potentially trapping the search in suboptimal regions. Moreover, $Q(a\mid s)$ estimates are often unreliable during early training, rendering the model highly susceptible to estimation errors. The linearly rescaled $Q(a\mid s)$ distribution thus provides a soft selection mechanism, facilitating a more nuanced balance between exploitation and exploration.
The hyperparameter $\alpha$ governs the relative weight of $p$, effectively controlling the degree of greediness in action selection. Higher values of $\alpha$ favor greedier behavior, emphasizing the $Q(a\mid s)$-informed policy, whereas lower values promote a more exploratory stance. In the limiting case where $Q(a\mid s)$ is uniform, Eq.~(\ref{eq: categorical}) reduces to sampling purely according to $P_F$.

In summary, our framework allows adaptive modulation of exploration and exploitation during PUCT-guided selection phase. By tuning $\alpha$, one can precisely control the model’s inclination toward greediness, concurrently accounting for both the global flow network and the local value distribution, thereby rendering the search process both flexible and controllable. The pseudocode is provided in Sec.~\ref{sec: Detailed Algorithm}.
\vspace{-0.3cm}
\section{Experiments}
\vspace{-0.1cm}
In this section, we evaluate the performance of \modelname on two tasks: \textbf{Hypergrid} and \textbf{Molecule Design}. These tasks are designed to test the model under different conditions: the former involves long action trajectories with sparse rewards, while the latter involves short trajectories with a large action space, also under sparse rewards. Our evaluation focuses on three central research questions (\textbf{RQ}): 
\vspace{-0.4cm}
\begin{itemize}[leftmargin=3em,itemsep=-0.1em]
\item[\textbf{RQ \ding{172}}] How quickly can the model discover high-reward candidates during the early stages of training?  
\item[\textbf{RQ} \ding{173}] How effectively can the model sustain the generation of high-reward candidates over time?  
\item[\textbf{RQ} \ding{174}] How well does the model maintain diversity among the generated candidates?  
\end{itemize}
\vspace{-0.2cm}

\begin{figure*}[t]
\centering
\begin{minipage}[t]{0.31\linewidth}
    \centering
    \vspace{-0.001\baselineskip} 
    \parbox{\linewidth}{
        \centering
        \captionof{table}{\small \textbf{Number of states visited to achieve the average reward of the top 100 candidates (lower is better).} \modelname\ (\textit{with} $Q$) denotes the full PUCT configuration, whereas \modelname\ (\textit{without} $Q$) corresponds to a PUCT variant without the greedy term. \textbf{Boldface} indicates the best-performing result.}
        \label{tab:average_top_100}
    }\vspace{-0.15cm}
    \scalebox{0.91}{
    \small
    \setlength{\tabcolsep}{3pt}
    \begin{tabular}{l|ccc}
    \toprule
    \multirow{2}{*}{\makecell[l]{\textbf{States}\\\textbf{visited}}} &
    \multicolumn{3}{c}{\textbf{Average top 100}} \\
    \cmidrule(lr){2-4}
    & \boldmath$>7.0$ & \boldmath$>7.5$ & \boldmath$>8.0$ \\
    \midrule
    TB        & 2,824 & 6,425 & 12,816 \\
    QGFN      & 2,000 & 2,800 & 10,800 \\
    \rowcolor{gray!15}
    \modelname (\textit{with} $Q$) & \textbf{644} & \textbf{964} & \textbf{5,204} \\
    \rowcolor{gray!15}
    \modelname (\textit{without} $Q$) & 1,524 & 2,404 & 9,204 \\
    \bottomrule
    \end{tabular}}
\end{minipage}
\hfill
\begin{minipage}[t]{0.65\linewidth}
    \centering
    \vspace{-0.3\baselineskip}
    \includegraphics[width=0.98\linewidth]{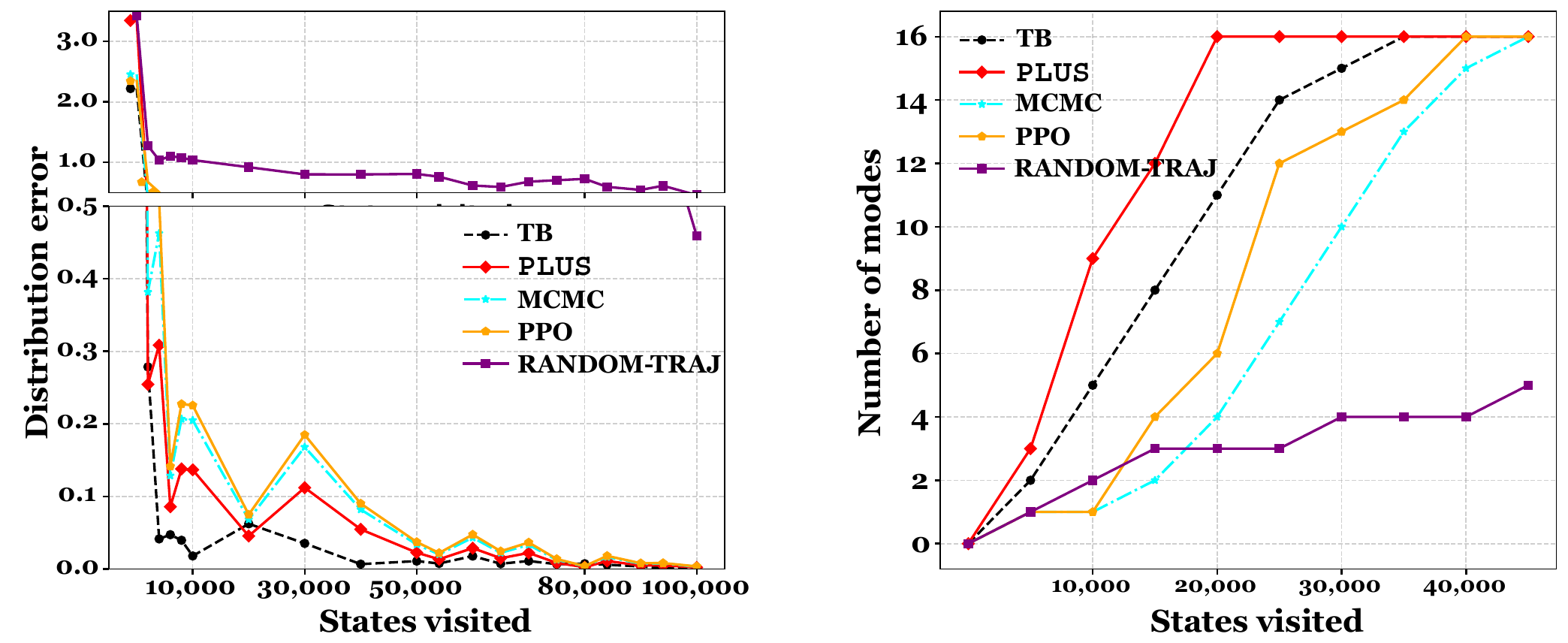}
    \vspace{-0.2\baselineskip}
    \parbox{\linewidth}{
        \centering
        \vspace{-0.15cm}
        \caption{\textbf{High-reward mode discovery and distribution matching on Hypergrid.}
\textit{Left}: $\ell_1$ loss between the learned sampling distribution and the target reward distribution. \textit{Right}: Number of high-reward region modes discovered with the same number of visits.}
\label{fig:hypergrid_modes_and_error}
    }
\end{minipage}
\vspace{-0.3cm}
\end{figure*}

\vspace{-0.3cm}
\subsection{Hypergrid Task}
\vspace{-0.1cm}
\paragraph{Task Description.}
We begin our evaluation with the Hypergrid environment introduced by~\citet{bengio2021flow}, a canonical testbed for assessing compositional generalization in GFlowNets. The environment consists of a $D$-dimensional discrete state space structured as a hypercube with edge length $H$, yielding $H^D$ distinct states. This task challenges agents to develop long-horizon planning capabilities while learning from extremely sparse reward signals.
The agent initiates each episode at the origin $(0,0,\cdots, 0) \in \mathbb{Z}^D$ and executes actions by incrementing any single coordinate by 1 (\textit{i.e.}, $\Delta x_d = 1$ for dimension $d$). From any state, the agent may alternatively choose a termination action that yields a reward determined by the following function:
\vspace{-0.2cm}
\begin{equation}
\small
\label{eq:reward}
\begin{aligned}
    R(\mathbf{x}) = R_0 + R_1 \prod_{d=1}^D \mathbb{I}\left(\left|\frac{x_d}{H-1}-0.5\right| \in (0.25, 0.5]\right) 
                \\+ R_2 \prod_{d=1}^D \mathbb{I}\left(\left|\frac{x_d}{H-1}-0.5\right| \in (0.3, 0.4]\right),
\end{aligned}
\vspace{-0.2cm}
\end{equation}
where $\mathbb{I}$ denotes the indicator function, and we adopt the standard parameterization: $R_0=10^{-5}$, $R_1=0.5$, $R_2=2$, with grid parameters $H=8$, $D=4$. 
\vspace{-0.3cm}
\paragraph{Metrics.}
Since the grid environment is relatively simple with only 16 modes, we adopt the following two metrics to evaluate the performance of our model: 1) \textbf{Number of modes}, which reflects the model's exploration capacity and structural diversity. 2) \textbf{The $\ell_1$ error} $\mathbb{E}_{x\sim p}\left[\left|p(x) - \frac{\mathcal{R}(x)}{Z}\right|\right]$, where $Z = \sum_x \mathcal{R}(x)$, measuring how well the learned sampling distribution $p(x)$ matches the target reward distribution. This $\ell_1$ error directly assesses whether the GFlowNets achieve their fundamental objective of generating samples with probabilities proportional to their rewards.
\vspace{-0.3cm}
\paragraph{Baselines.}
We compare \modelname with representative flow-based baselines like TB and MCMC~\citep{malkin2022trajectory, bengio2021flow, zhang2022generative}, as well as non-flow-based methods, including PPO~\citep{schulman2017proximal} and RANDOM-TRAJ (which samples actions uniformly at random). All methods are evaluated under the same grid environment and reward function to ensure fairness. 

\vspace{-0.3cm}
\paragraph{Early Discovery of High-Reward Candidates (RQ \ding{172}).}  
The right panel of Figure~\ref{fig:hypergrid_modes_and_error} shows the number of high-reward modes discovered as a function of the number of state visits. \modelname reaches 8 modes after approximately 10,000 visits, whereas TB requires around 15,000 visits to reach the same count. MCMC and PPO reach 8 modes later, and RANDOM-TRAJ performs the slowest. These results indicate that \modelname allocates sampling effort more efficiently toward high-reward regions early on, leading to faster accumulation of high-reward modes compared to baselines.

\vspace{-0.3cm}
\paragraph{Sustained Generation of High-Reward Candidates (RQ \ding{173}).}  
As the number of visits increases, \modelname continues to generate high-reward samples effectively, eventually recovering all 16 modes with fewer visits than baselines. In contrast, RANDOM-TRAJ and other baselines accumulate high-reward modes more slowly, reflecting a more uniform sampling across the state space. This pattern shows that \modelname consistently revisits high-reward states identified earlier, allowing the sampling process to maintain a higher concentration of high-reward candidates over time without explicit redistribution.
\vspace{-0.3cm}
\paragraph{Diversity of Generated Candidates. (RQ \ding{174})}  
The left panel of Figure~\ref{fig:hypergrid_modes_and_error} reports the $\ell_1$ distribution error, quantifying the deviation between the learned sampling distribution and the reward-proportional objective. Vanilla GFlowNets achieve lower $\ell_1$ error due to strict adherence to proportionality, resulting in broad coverage across all modes. \modelname exhibits moderately higher $\ell_1$ error because its MCTS-guided strategy allocates more sampling to high-reward states. Because our goal is not to strictly approximate the full distribution, but to efficiently generate more high-reward candidates within limited sampling resources, while retaining sufficient coverage to avoid collapsing diversity. RANDOM-TRAJ, by comparison, shows both high $\ell_1$ error and poor mode coverage, highlighting the impact of guided sampling on achieving both efficiency and balanced diversity.

\vspace{-0.3cm}
\subsection{Molecule Design Task}
\vspace{-0.1cm}
\paragraph{Task Description.}
Recent advances in artificial intelligence have revolutionized computational chemistry, particularly in molecular property prediction and design~\citep{du2024mmgnn,li2024molclw,zhang2023improving}. Molecular design presents an ideal application scenario for GFlowNets, as it requires simultaneous optimization of two critical objectives: 1) \emph{quality} (achieving target chemical properties) and 2) \emph{diversity} (generating structurally distinct candidates). This dual requirement stems from practical drug discovery needs, where viable candidates must not only exhibit strong binding affinities but also possess synthesizable structures. We focus on the specific challenge of designing molecules with maximal binding energy to a target protein. To this end, we formally describe the action space for molecular generation: building on junction tree-based molecular generation and following~\citet{bengio2021flow}, we define:
\vspace{-0.2cm}
\begin{equation}
\small
\mathcal{A}(s) = \{(v,b) \mid v \in \mathcal{V}(s), b \in \mathcal{B}\},
\vspace{-0.2cm}
\end{equation}
where $v$ denotes the choice of target atom, $b$ denotes the choice of building block.
where $\mathcal{V}(s)$ denotes attachable atoms in state $s$ and $\mathcal{B}$ is our building block vocabulary ($|\mathcal{B}|=105$). Given a molecule, a building block can be added to the molecule at different positions. The combinatorial action space poses significant exploration challenges while enabling the generation of diverse molecular scaffolds. In our experiments, we adhere closely to the setup of~\citet{bengio2021flow}, including the construction of the action space, the use of the junction tree-based molecular representation, and the deployment of a reward model for evaluating candidate molecules. By following this established protocol, we ensure that our evaluation is directly comparable and grounded in prior work, while focusing on the efficiency and diversity of high-reward molecule generation.

\begin{figure*}[t]
\centering
\begin{minipage}[t]{0.655\linewidth}
    \centering
    \includegraphics[width=\linewidth]{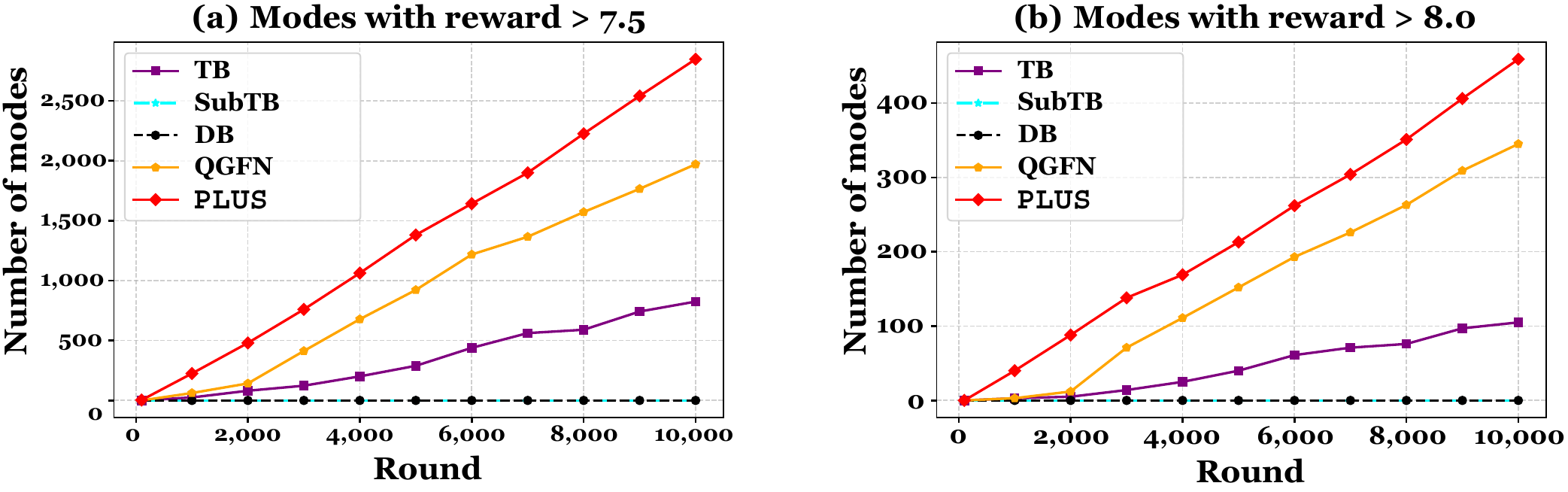}
    \caption{\textbf{Number of modes with reward $>7.5$ and $>8.0$ across different models in molecule design task.} \textit{Left}: \textbf{(a)} Model comparison of number of modes with reward $>7.5$. \textit{Right}: \textbf{(b)} Model comparison of umber of modes with reward $>8.0$.}
    \label{fig:n_mode_7.5}
\end{minipage}
\hfill
\begin{minipage}[t]{0.315\linewidth}
    \centering
    \includegraphics[width=\linewidth]{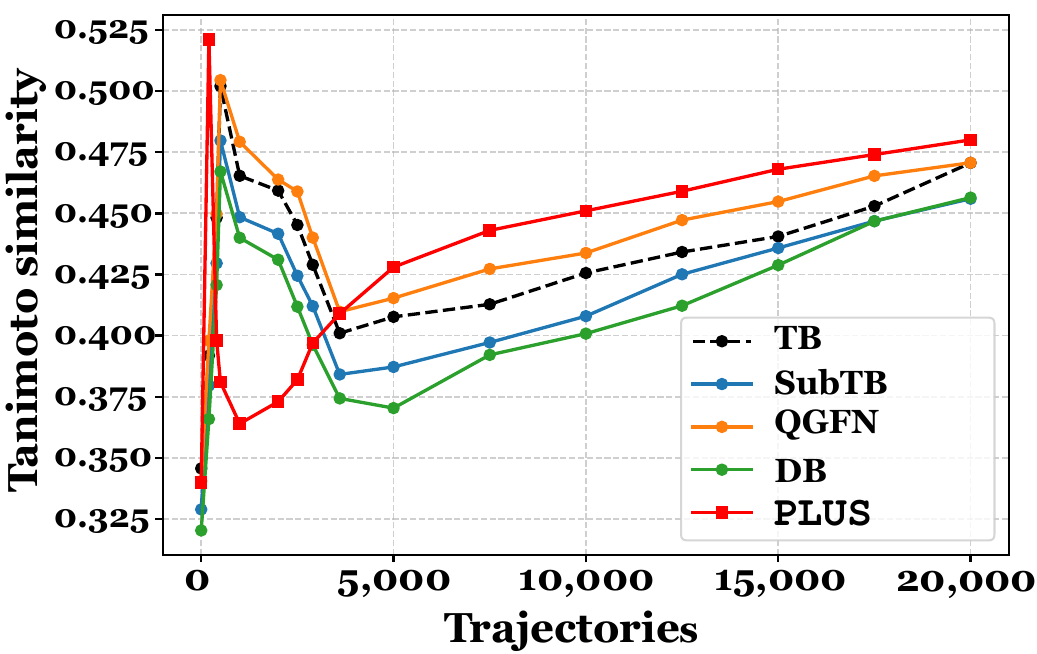}
    \caption{\textbf{Tanimoto similarity} among the top 1,000 highest-reward molecules generated by different models.} 
    \label{fig: tanimoto}
\end{minipage}
\vspace{-0.3cm}
\end{figure*}

\vspace{-0.3cm}
\paragraph{Metrics.}

1) \textbf{Number of modes}, which reflects the model’s exploration capacity and structural diversity. In the molecule task, we define a \emph{mode} as a unique terminal molecule after canonicalization (canonical SMILES) and deduplication; when we report “modes with reward $>B$”, we count the number of unique terminal molecules whose reward exceeds the threshold $B$. 2) \textbf{Average top 100}, among all generated candidate molecules, we report how many molecular states are visited when the top-100 average reward exceeds 7.0, 7.5, and 8.0. Fewer visited states to reach the corresponding average reward indicate faster discovery of high reward regions. 
In molecular generation tasks, if all high-reward molecules produced are structurally almost identical, then even high reward values would indicate that the model suffers from mode collapse.
To address this, we additionally adopt the 3) \textbf{Tanimoto similarity} metric, which measures the structural differences among generated molecules and further reflects whether the model can maintain diversity while consistently generating high-reward molecules.
\vspace{-0.3cm}
\paragraph{Baselines.} 
we compare \modelname with four popular flow-based baselines, TB~\citep{malkin2022trajectory}, SubTB~\citep{madan2023learning}, DB~\citep{bengio2021flow}, and QGFN~\citep{lau2024qgfn}. Here, we adopt the same training parameters as the vanilla GFlowNets.
  
\vspace{-0.3cm}
\paragraph{Early-Stage Discovery of High-Reward Candidates (RQ \ding{172}).}
Figure~\ref{fig:n_mode_7.5} illustrates the efficiency of different models in identifying high-reward candidates during the training iterations. SubTB and DB exhibit limited capability, failing to locate any samples with rewards exceeding $7.5$ or $8.0$ even after 10,000 iterations. TB achieves modest improvement, yet is substantially outperformed by \modelname and QGFN in the rapid identification of high-reward candidates. Notably, \modelname consistently discovers samples with rewards above $8.0$ within only 300 iterations. This early advantage stems from the incorporation of the exploration term in Eq.~\ref{eq: puct_short}, which adaptively modulates the selection of actions to prioritize states likely to lead to high-reward candidates. In contrast, QGFN’s reliance solely on $Q$-values occasionally leads to overestimation of intermediate-reward regions, resulting in inefficient allocation of computational effort.
\vspace{-0.3cm}
\paragraph{Sustained Generation of High-Reward Candidates (RQ \ding{173}).}
Table~\ref{tab:average_top_100} further quantifies the models’ performance in consistently generating high-reward candidates over time. The number of states visited to achieve the top-100 candidates across reward thresholds ($7.0$, $7.5$, $8.0$) demonstrates the persistent advantage of \modelname. While QGFN improves over TB through action-value guidance, inaccuracies in early-stage $Q$-estimates can divert the sampling process toward suboptimal trajectories. By integrating MCTS with an $\alpha$-greedy strategy, \modelname maintains robust performance across all thresholds, efficiently allocating sampling resources to regions associated with higher rewards. These results highlight that combining exploration-driven selection with learned value estimates stabilizes the generation of high-reward candidates throughout training, allowing the model to leverage accumulated experience effectively.

\vspace{-0.3cm}
\paragraph{Diversity of Generated Candidates (RQ \ding{174}).}
The Tanimoto similarity computed over the first 1,000 generated molecules (illustrated in Figure~\ref{fig: tanimoto}) serves as an indicator of structural diversity. Owing to the inclusion of more greedy components in the sampling procedure, \modelname exhibits the highest inter-molecular similarity among the compared methods. However, the final converged similarity is only marginally higher, by approximately $5\%$, than that achieved by QGFN, which adopts a comparable greedy sampling strategy. This observation suggests that the increased emphasis on greediness does not lead to a substantial degradation in diversity.
The moderate rise in similarity can be attributed to a more focused exploitation of high-reward regions in the search space, while the exploration term continues to promote traversal into less-explored areas of the molecular action space. This balance enables the model to uncover additional structural patterns without compromising reward efficiency. Consequently, \modelname is able to identify high-reward candidate molecules both rapidly and consistently, without incurring a pronounced loss in diversity.
It is worth emphasizing that this sampling strategy is not designed to faithfully reproduce the complete target distribution. Rather, it is motivated by the practical objective of efficiently discovering high-reward candidate molecules as early and as plentifully as possible.

\begin{table}[t]
\centering
\caption{\small \textbf{Number of modes (averaged over candidates with reward $> 8.0$) identified under different $\alpha$ settings at various training stages.} Boldface denotes the best result in each column. Temp indicates that $\alpha$ is linearly annealed from $0$ to $0.2$ over the course of training.}
\vspace{-0.15cm}
\small
\footnotesize
\begin{tabular}{l|ccc}
\toprule
{\textbf{Model settings}} 
& \textbf{12,000} & \textbf{24,000} & \textbf{40,000} \\
\midrule
\modelname($ \alpha=0.6$)   & 8  & 14  & 101 \\
\modelname($ \alpha=0.4$)   & 17  & 46  & 249 \\
\rowcolor{gray!15}
\modelname($ \alpha=0.2$)   & \textbf{92}  & \textbf{175} & \textbf{459} \\
\modelname($ \alpha=0$)     & 14  & 61  & 105 \\
\modelname($ Temp $)        & 1   & 26  & 103 \\
\bottomrule
\end{tabular}
\label{tab: ablation_avg_8.0}
\vspace{-0.4cm}
\end{table}

\vspace{-0.3cm}
\subsection{Sampling Efficiency of Different MCTS Variants}
\vspace{-0.15cm}
We conduct controlled ablation studies to isolate the contribution of MCTS in our framework and examine how varying the influence of MCTS-derived action values affects sampling efficiency. Table~\ref{tab: ablation_avg_8.0} reports results for different values of the greediness coefficient $\alpha$, which balances MCTS guidance against the GFlowNets forward policy. We also evaluate a temperature-annealed strategy, where $\alpha$ increases linearly during training to encourage early exploration and progressively favor exploitation.
\vspace{-0.3cm}
\paragraph{Effect of MCTS Guidance on Sampling Efficiency.}
To isolate MCTS impact, we compare $\alpha=0$, effectively a vanilla GFlowNet, against $\alpha>0$. As Table~\ref{tab: ablation_avg_8.0} shows, the $\alpha=0$ baseline discovers substantially fewer high-reward modes than $\alpha=0.2$ or $\alpha=0.4$ throughout training, demonstrating that MCTS guidance significantly enhances the identification and exploitation of high-reward regions. Interestingly, the temperature-annealed setting \textit{Temp} underperforms even $\alpha=0$ in early and mid training. We attribute this to an initially near-zero $\alpha$ combined with an unstable flow network: the model behaves almost like vanilla GFlowNet, while PUCT introduces a strong exploration bonus, resulting in excessive exploration and degraded sampling efficiency. These results underscore that explicit, well-calibrated MCTS guidance is crucial.
\vspace{-0.3cm}
\paragraph{Impact of Different Greedy Degrees on Sampling Efficiency.}
We next examine how $\alpha$ magnitude affects performance. Although larger $\alpha$ values—favoring exploitation—might be expected to boost early performance, higher $\alpha$ (e.g., $0.4$ and $0.6$) consistently underperform the moderate $\alpha=0.2$, even early in training. Excessive greediness leads to premature convergence on locally attractive but globally suboptimal regions, which may initially appear high-reward but ultimately yield low-reward trajectories. This effect intensifies in later training, where $\alpha=0.2$ discovers over four times as many high-reward modes as the $\alpha=0$ baseline at 40,000 steps.
Temperature-based annealing is intended to smooth the exploration–exploitation trade-off, yet its performance is highly sensitive to the schedule. Initially, small $\alpha$ values negate MCTS guidance, while PUCT exploration dominates early training, compounding over-exploration. Optimizing the annealing schedule is orthogonal to our main contributions, so we do not pursue it further.
Empirically, a fixed $\alpha=0.2$ provides the most effective and robust balance between exploration and exploitation, yielding consistently superior sampling efficiency across all training stages.

\begin{table}[t] 
    \centering
        \caption{\small \textbf{Number of states visited to discover different modes in the HyperGrid task.} We compare the performance of the PUCT-based selection strategy, a random selection strategy, and a variant that omits the exploration term during the MCTS selection phase. 
        }
        \vspace{-0.15cm}
    \small
    \begin{tabular}{l|ccc}
        \toprule
        \textbf{States visited} & \textbf{4} & \textbf{8} & \textbf{16} \\
        \midrule
        \modelname (\textit{Random}) & 5,734 & 13,645 & 30,754 \\
        \rowcolor{gray!15}
        \modelname (\textit{PUCT}) & \textbf{4,816} & \textbf{9,616} & \textbf{20,816} \\
        \modelname (\textit{PUCT Without Explore}) & 6,416 & 19,216 & 49,616 \\
        \bottomrule
    \end{tabular}
    \label{tab: puct vs random policy in selection stage}
    \vspace{-0.4cm}
\end{table}

\vspace{-0.3cm}
\subsection{Ablation of Selection Strategies: PUCT vs. Random}
\vspace{-0.1cm}
To isolate the effect of the selection strategy in the selection stage within MCTS, we compare PUCT against a random policy while keeping all other components fixed. As reported in Table~\ref{tab: puct vs random policy in selection stage}, PUCT consistently discovers the same number of modes with substantially fewer state visits: for $4$, $8$, and $16$ modes, the visited states decrease from $5{,}734$ to $4{,}816$, $13{,}645$ to $9{,}616$, and $30{,}754$ to $20{,}816$, respectively.
The advantage of PUCT grows with problem complexity, reflecting improved scalability. Early in training, PUCT promotes broad but targeted exploration guided by policy priors, gradually shifting to focused exploitation as value estimates stabilize. By contrast, random selection disperses computation across largely uninformative branches. These results underscore the critical role of principled exploration–exploitation balancing in the selection stage and highlight PUCT as a key driver of the efficiency of our MCTS-guided sampler.
\vspace{-0.3cm}
\subsection{Roles of Exploration and Exploitation in PUCT-Based Selection}
\vspace{-0.2cm}
We investigate how value-guided greediness shapes sampling dynamics on the molecular design task. Performance is measured by the average reward of the top-100 generated samples, comparing policy variants with and without value exploitation. As shown in Table~\ref{tab:average_top_100}, achieving the same reward thresholds without the greedy term requires substantially more state visits, indicating that high-reward candidates are encountered less frequently. Incorporating greediness consistently improves sampling efficiency by directing search toward high-reward regions once informative value estimates are available, while maintaining sufficient exploration in earlier stages. In contrast, removing value exploitation disperses search effort across low-reward regions, preventing reliable revisitation of high-quality candidates. These results highlight controlled greediness as essential for sustaining high-reward generation without prematurely collapsing exploration.

To examine the role of explicit exploration, we consider the Hypergrid task and compare variants with and without the exploration term. As Table~\ref{tab: puct vs random policy in selection stage} shows, omitting this term dramatically increases the number of state visits required to discover the same number of modes, with performance degrading by over a factor of two for larger mode counts. This effect intensifies as the search problem becomes more challenging. Structured exploration is particularly important early in training, when value estimates are noisy: without it, the model overcommits to spurious $Q$-value fluctuations, biasing the search toward poorly supported regions. Including the exploration term ensures broader yet informed coverage early on, allowing value estimates to stabilize before exploitation dominates. Consequently, sampling progresses more effectively from exploration to exploitation, yielding substantially improved efficiency and robustness.

\begin{table}[t]
\centering
\caption{\small \textbf{Number of modes identified under two reward thresholds for different $c_{puct}$.} \textbf{Boldface} indicates the best results. 
}
\vspace{-0.15cm}
\resizebox{\linewidth}{!}{
\begin{tabular}{l|ccc|ccc}
\toprule
\multirow{2}{*}{\textbf{Configuration}}
& \multicolumn{3}{c|}{\textbf{average top $>7.5$}}
& \multicolumn{3}{c}{\textbf{average top $>8.0$}} \\
\cmidrule(lr){2-4}\cmidrule(lr){5-7}
& \textbf{12,000} & \textbf{24,000} & \textbf{40,000}
& \textbf{12,000} & \textbf{24,000} & \textbf{40,000} \\
\midrule
\modelname ($c_{\mathrm{puct}}=0.25$)
& 279 & 457 & 864 & 78 & 143 & 254 \\
\modelname ($c_{\mathrm{puct}}=0.50$)
& 402 & 947 & 1,879 & 128 & 312 & 704 \\
\modelname ($c_{\mathrm{puct}}=1.00$)
& 507 & \textbf{1,080} & \textbf{2,848} & 165 & \textbf{384} & \textbf{1,053} \\
\modelname ($c_{\mathrm{puct}}=2.00$)
& \textbf{546} & 923 & 1,645 & \textbf{189} & 301 & 612 \\
\modelname ($c_{\mathrm{puct}}=4.00$)
& 325 & 634 & 1236 & 132 & 201 & 375 \\
\bottomrule
\end{tabular}}
\label{tab: ablation_merge}
\vspace{-0.4cm}
\end{table}

\vspace{-0.3cm}
\subsection{Effect of Different Exploration Coefficient $c_{puct}$} 
\vspace{-0.1cm}
As shown in Table~\ref{tab: ablation_merge}, reducing the exploration coefficient \(c_{\text{puct}}\) substantially impairs the model’s ability to consistently generate high-reward samples. This aligns with our hypothesis: smaller \(c_{\text{puct}}\) restricts exploration of less-visited regions, causing the model to miss high-reward areas during early training. Conversely, larger \(c_{\text{puct}}\) enhances early performance by promoting broader exploration and facilitating the initial discovery of high-reward trajectories. However, excessive exploration eventually yields diminishing returns, as the PUCT selection formula (Eq.~\ref{eq: puct_short}) becomes dominated by the exploration term, leading to over-exploration and suboptimal convergence. Through extensive empirical evaluation, we find that \(c_{\text{puct}} = 1\) strikes an effective balance between exploration and exploitation, enabling early discovery of high-reward regions while allowing accumulated experience to guide sustained generation of top-performing samples.

\vspace{-0.3cm}
\section{Conclusion}
\vspace{-0.15cm}
In this paper, we address the challenge of high-reward discovery in structured generation by focusing on settings where early guidance and sustained performance are essential. We propose \modelname, a planning-augmented sampling framework that integrates guided selection with a controllable soft-greedy mechanism, enabling targeted early exploration and a smooth transition toward exploitation as experience accumulates. Comprehensive experimental results demonstrate that our approach improves sampling efficiency and consistently sustains high-reward generation, highlighting planning-augmented sampling as an effective paradigm for early high-reward discovery in generation tasks.

\newpage

\section*{Impact Statements}
This paper presents work whose primary goal is to advance the field of machine learning, particularly in the area of efficient sampling for structured generation. While such advances may have downstream applications in a variety of domains, the ethical considerations and broader societal impacts are those commonly associated with general progress in machine learning research. We do not identify any specific societal consequences that require separate discussion beyond this context.

\bibliography{example_paper}
\bibliographystyle{icml2026}

\appendix

\section*{Appendix Contents}
\vspace{0.2cm}
\hrule
\textbf{A. Background for Monte Carlo Tree Search} \dotfill Page~\pageref{sec: monte carlo tree search}\\
     \textbf{B. Pseudocode of Framework} \dotfill Page~\pageref{sec: Detailed Algorithm}\\
     \textbf{C. Challenge in Backpropagation} \dotfill Page~\pageref{sec: Backpropagation Challenge Details}\\
     \textbf{D. More Experimental Details} \dotfill Page~\pageref{sec: Additional Experimental Results}\\
     \textbf{E. Discussion} \dotfill Page~\pageref{sec:discuss}\\
     \vspace{-0.1cm}
\hrule

\begin{algorithm}[t]
\small
\caption{Action sampling process of \modelname.}
\label{alg:algorithm}
\begin{algorithmic}[1]
\State \textbf{Input:} Reward function $R: \mathcal{X} \to \mathbb{R}_{>0}$, batch size $M$, model $P_F$ with parameters $\theta$, initialized state $s_0$, number of MCTS iterations $I$, PUCT exploration coefficient $c_{\text{puct}}$, greediness factor $\alpha$, current state $s_i$
\State \textbf{Output:} MCTS sampling policy
\Statex \hspace{-\algorithmicindent} \textit{// $I$ iterations}
\For{$i = 1$ to $I$}
\Statex \hspace{-\algorithmicindent} \textit{// Selection stage}
    \State \textbf{Selection:} select state from $s_i$ to $s_j$ by selection strategy; record path $\tau = (s_0 \to \cdots \to s_i \to \cdots \to s_j)$
    \If{$s_j$ is a terminal state}
        \State \textbf{Backpropagation:} Propagate reward $\mathcal{R}(s_j)$ along the trajectory $\tau$
    \Else
        \Statex \hspace{-\algorithmicindent} \textit{// Expansion stage}
        \State \textbf{Expansion:} initialize the $Q(a\mid s_j)$ and $N(a\mid s_j)$ of all action belong to $\mathcal{A}(s_j)$
        \Statex \hspace{-\algorithmicindent} \textit{// Simulation stage}
        \State \textbf{Simulation:} Choose one action from the initialized action during the expansion stage, and roll out to the terminal state $x$ using $P_F$
        \Statex \hspace{-\algorithmicindent} \textit{// Backpropagation stage}
        \State \textbf{Backpropagation:} Propagate reward $\mathcal{R}(x)$ along $\tau$
    \EndIf
\EndFor
\Statex \hspace{-\algorithmicindent} \textit{// Mixing probability sampling}
\State \textbf{Sampling action:} Use $\alpha$-greedy over $Q(a\mid s_i)$ predicted and $P_F$ to select action to execute.
\end{algorithmic}
\end{algorithm}

\section{Background for Monte Carlo Tree Search} 
\label{sec: monte carlo tree search}
Monte Carlo Tree Search (MCTS) \citep{coulom2006efficient} is a best-first search algorithm that combines tree search with Monte Carlo simulation. The algorithm iteratively builds a search tree through four key phases:
$\text{Selection} \rightarrow \text{Expansion} \rightarrow \text{Simulation} \rightarrow \text{Backpropagation}$.

\begin{itemize}
    \item \textbf{Selection}: Traverse the tree from root to leaf using a tree policy (typically Upper Confidence Bound for Trees, UCT)~\citep{kocsis2006bandit}:
    \begin{equation}
    a^* = \underset{a}{\text{argmax}} \left( Q(a\mid s) + c \sqrt{\frac{\ln N(s)}{N(a\mid s)}} \right),
    \label{eq:uct}
    \end{equation}
    where $Q(a\mid s)$ is the action value, $N(s)$ and $N(a\mid s)$ are visit counts, and $c$ is an exploration constant.

    \item \textbf{Expansion}: When reaching an expandable node, create one or more child nodes representing possible state transitions.

    \item \textbf{Simulation}: Perform a Monte Carlo rollout from the expanded node using a default policy to estimate the reward.

    \item \textbf{Backpropagation}: Update statistics along the traversed path:
\end{itemize}

\begin{figure*}[t]
    \centering
    \includegraphics[width=0.7\linewidth]{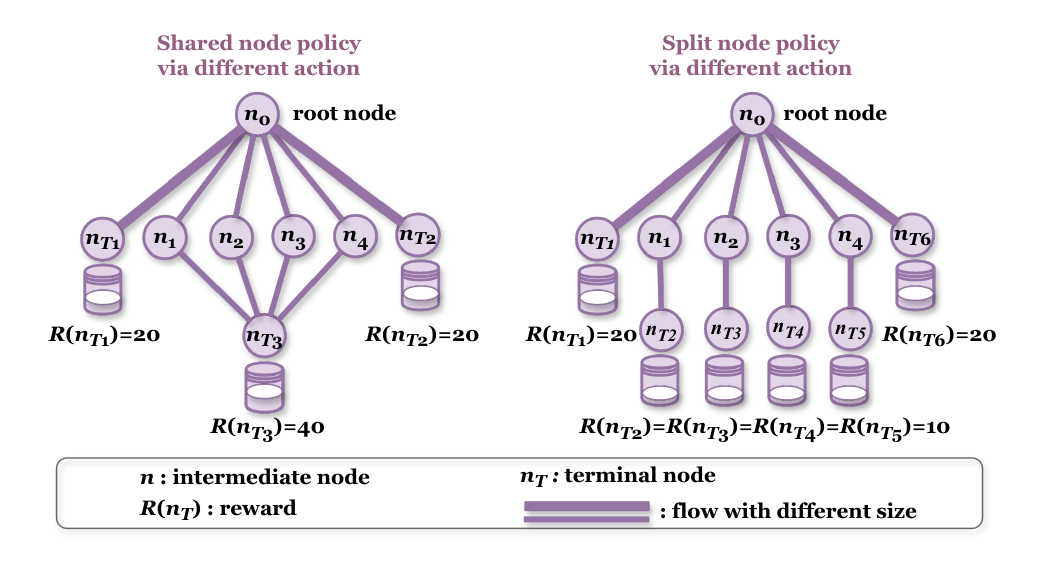}
    \caption{\textbf{Comparison of different representations for reaching the same state via multiple action sequences}. On the \textit{left}, identical states are represented by a single shared node; On the \textit{right}, the same state reached through different action sequences is represented by distinct nodes.}
    \label{fig: different actions}
    \vspace{-0.2cm}
\end{figure*}

\section{Pseudocode of Framework}
\label{sec: Detailed Algorithm}
This section describes the detailed algorithmic flow of our framework, as shown in the Algorithm~\ref{alg:algorithm}.

\section{Challenge in Backpropagation} 
\label{sec: Backpropagation Challenge Details}
There is a challenge in updating the nodes during the backpropagation phase. Because updating different parents leads to a completely different distribution of the flow network. \textbf{There are some alternative options:} 1) The reward $\mathcal{R}(x)$ is uniformly propagated back to each parent node, such that if there are $n$ parent nodes, each parent node updates its own $Q$-value with $\mathcal{R}(x)/n$. 2) Distribute the reward $\mathcal{R}(x)$ to all parent nodes proportionally based on their relative flow magnitudes within the flow network. Each parent node updates its own $Q$-value with $\rho \mathcal{R}(x)$, where $\rho$ represents the proportion of flow from the parent node to a specific child node. 3) the reward $R(x)$ is only propagated back along the trajectory $\tau = ({s}_{0}\rightarrow ...\rightarrow{s}_{i})$ in the selection phase, described in detail in Sec.~\ref{sec: Targeted Selection of high-reward Regions via PUCT}. Each node included in the trajectory $\tau$ updates its own $Q$-value with $\mathcal{R}(x)$. 

In our work, we adopt the third method for the following reasons. The first two approaches require updating all parent nodes and iteratively propagating these updates further up the tree by updating the parents of parents and so on. This \textbf{results in significant computational overhead}. Moreover, the second method incurs additional cost by computing the proportion of flow from each parent to its children, which further increases the computational burden.
Furthermore, \textbf{restricting updates to nodes along the selected path $\tau$ serves to emphasize the highest-reward trajectory}. In contrast, the first two methods would dilute the relative contribution of this highest-reward path by distributing credit more broadly, which is undesirable. For different actions that lead to the same node, we have designed a global mapping of state nodes. For identical states, only one node is preserved. This approach also aligns with the objective of flow network training. As shown in Figure~\ref{fig: different actions}, \textbf{if we create multiple nodes for the same terminal state via different action orders, we will distribute the proportion of high reward regions among these nodes, which may prevent the MCTS tree from accurately reflecting the high reward characteristics of these terminal nodes.}
Given these considerations, we opt for the third approach in our experimental design.

\section{More Experimental Details}
\label{sec: Additional Experimental Results}

\subsection{Additional Experimental Results of Comparison of Different Expanding Strategies}
In the main text, we mentioned that our expanding strategy is adding all child nodes to the unexpanded node, because the exploration term in our PUCT formulation (Eq.~(\ref{eq: puct_short})) guarantees that these newly created nodes will be properly prioritized based on their low $n_{visit}$, ensuring they will be systematically explored in future iterations.
To better validate the rationality of our design, we conducted a comparative experiment in the Hypergrid environment, comparing the approach of expanding all child nodes versus expanding only one child node. \textit{The goal was to measure the number of states visited required to discover the same number of modes}. If discovering the same modes requires visiting significantly more states, it indicates wasted MCTS iterations and lower state-visit efficiency, making such an approach less desirable. The results of this comparative experiment are shown below:

\begin{table}
\centering
\caption{\textbf{The number of states visited for different modes discovered}. Results of comparing the approach of expanding all child nodes versus expanding only one child node, fewer states visited to discover the same number of modes, indicate higher exploration efficiency.}
\small
\begin{tabular}{l|c|c|c}
    \toprule
    \textbf{States visited} & \textbf{4} & \textbf{8} & \textbf{16} \\
    \midrule
    \modelname (\textit{expand all}) & \textbf{4,816} & \textbf{9,616} & \textbf{20,816} \\
    \modelname (\textit{expand one}) & 9,616 & 134,416 & / \\
    \bottomrule
\end{tabular}
\label{tab: comparison between adding all child and adding one child}
\end{table}

Since the strategy of expanding only one child node requires an impractically large number of state visits to discover all 16 modes (rendering it meaningless for comparison), we omit this result here. However, the state visit counts required for discovering 4 and 8 modes clearly demonstrate the infeasibility of single child expansion. Our results show that this approach significantly reduces exploration efficiency.
We attribute this inefficiency to the fundamental limitation of single child expansion: \textbf{Each training iteration predominantly revisits previously explored nodes due to constrained graph width in MCTS. This severe restriction on new node access dramatically reduces the exploration space.} Even in our grid experiment, the state visit counts reached alarming magnitudes, let alone in molecular experiments with exponentially larger state spaces, where such costs would become computationally prohibitive.
These experimental results conclusively validate our design rationale: expanding all valid child nodes during the expansion phase is essential for achieving optimal state visitation and exploration efficiency.

\subsection{Detailed Experimental Setup}
\paragraph{Parameter Setup in Hypergrid Task.}
For the GFlowNets policy model, we use the same configuration as vanilla GFlowNets, 
and we sampled trajectories with a batch size of 16, using the Adam optimizer
with all other parameters at their default values. All experiments in this task are performed on a CPU. The horizon and dimension are set to 8 and 4.
For the MCTS framework, we set the maximum depth of simulation to 20, the exploration coefficient to 1, and the greediness factor to $0.2$. 

\vspace{-0.2cm}
\paragraph{Parameter Setup in Molecule Design Task.}
For the GFlowNets policy model, we use the dataset and proxy model provided by ~\citet{bengio2021flow, lau2024qgfn, malkin2022trajectory}. Different from the hypergrid experiment, due to the large state space of this experiment, we set the maximum depth of simulation to 8, the exploration coefficient to 1, and the greediness factor to $0.2$.

\section{Discussion}
\label{sec:discuss}
\subsection{When to Use \modelname}
Our proposed \modelname is most suitable for structured generation settings where the primary objective is \emph{efficient early discovery} and \emph{sustained generation} of high-reward candidates, rather than faithful matching of the full reward-induced distribution. 
Such regimes commonly arise in settings where early identification of high-quality candidates is particularly valuable.
In these settings, approaches that emphasize global distribution matching or broad mode coverage may allocate substantial effort to low-reward regions during early training, while purely greedy value-driven strategies can be sensitive to unreliable value estimates at initial stages. 
By integrating planning-guided selection with a controllable soft-greedy mechanism, \modelname provides a complementary trade-off: it enables targeted early guidance and supports a gradual transition toward exploitation as experience accumulates, without committing to a single rigid decision rule. 
This perspective also helps clarify why not all baselines are directly aligned with the same objective, as methods primarily optimized for distributional fidelity or late-stage exploitation address operating regimes that differ from the early-discovery focus studied here.

\subsection{Limitations}
Our study considers controlled environments with fixed action spaces and stationary reward functions, which provide a clear and consistent setting for validating the core ideas of planning-augmented sampling and for conducting systematic comparisons across methods. 
While this scope captures the target regime studied in this work, it does not explicitly cover scenarios with dynamically evolving action sets or nonstationary reward distributions. 
In addition, our evaluation focuses on sample efficiency, measured by the number of state visits required to discover high-reward solutions, rather than wall-clock runtime; a detailed analysis of runtime trade-offs therefore falls outside the scope of this paper. 
Finally, our method is designed to prioritize efficient high-reward discovery, and applications that require accurate reward-proportional sampling over the entire solution space may favor alternative distribution-matching approaches.

\subsection{Future Work}
Future work may extend \modelname to broader generation settings beyond the controlled environments studied here. 
One natural direction is to support more dynamic problem formulations, such as evolving action sets or changing constraints, while preserving the intended exploration--exploitation behavior of planning-augmented sampling. 
Another promising avenue is to develop more adaptive mechanisms for balancing exploration and exploitation, for example, by adjusting key hyperparameters based on the stability of value estimates or observed search dynamics. 
In addition, further improving the scalability of planning-augmented sampling, such as through more selective expansion strategies or reuse of search statistics, could facilitate its application to larger and more complex generation problems.

\end{document}